\documentclass{article}

\usepackage{PRIMEarxiv}

\usepackage[utf8]{inputenc} % allow utf-8 input
\usepackage[T1]{fontenc}    % use 8-bit T1 fonts

\usepackage{graphicx}
\usepackage{subcaption}
\usepackage{algorithm}
\usepackage{color}
\usepackage{algpseudocode}
\usepackage{amsmath}
\usepackage{amsthm}
\usepackage{multirow}
\usepackage{bm}
\newtheorem{theorem}{Theorem}

\newtheorem{definition}{Definition}
\usepackage{amsfonts}
\usepackage{booktabs} % for professional tables

\usepackage{hyperref}       % hyperlinks
\title{Improving Training-free Conditional Diffusion Model via Fisher Information}

\author{
  Kaiyu Song, Hanjiang Lai \\
  Sun Yat-Sen University \\
  \texttt{\{songky7, laihanj3\}@mail2.sysu.edu.cn}
}
\begin{document}
\maketitle

\begin{abstract}
Training-free conditional diffusion models have received great attention in conditional image generation tasks. 
However, they require a computationally expensive conditional score estimator to let the intermediate results of each step in the reverse process toward the condition, which causes slow conditional generation.  
In this paper, we propose a novel Fisher information-based conditional diffusion (FICD) model to generate high-quality samples according to the condition. In particular, we further explore the conditional term from the perspective of Fisher information, where we show Fisher information can act as a weight to measure the informativeness of the condition in each generation step. According to this new perspective, we can control and gain more information along the conditional direction in the generation space. Thus, we propose the upper bound of the Fisher information to reformulate the conditional term, which increases the information gain and decreases the time cost. Experimental results also demonstrate that the proposed FICD can offer up to 2x speed-ups under the same sampling steps as most baselines. Meanwhile, FICD can improve the generation quality in various tasks compared to the baselines with a low computation cost.
\end{abstract}    
\section{Introduction}

Recently, unconditional diffusion models (DMs)~\cite{ddpm,sde,ddim} have shown great success in image generation tasks. However, people often want to generate images with the properties they want. Conditional diffusion models~\cite{controlnet}, which incorporate the conditions to generate the desired properties, have emerged as a crucial role for various generation tasks, e.g., text-image generations~\cite{sd,glide,endtoend}, style-driven generation~\cite{pnp}, and image edit-driven generation~\cite{freedom,controlnet,sdedit}. Many researchers~\cite{dreambooth,freedom} have proposed to reuse/fine-tune the mature unconditional diffusion models to improve conditional generation tasks.

From unconditional to conditional diffusion models~\cite{dps}, the main problem is how to incorporate conditional information into the diffusion model to guide the sample generation. A new conditional score estimator ~\cite{diifusionbeatgan,sd} should be introduced into the diffusion models. However, it is not trivial to define the conditional score estimator. This is because the diffusion model is a multi-step generation. In each step, the intermediate-generated results are sampled from different noisy distributions. This hinders us from directly measuring the distance between the condition and intermediate results to generate desired images. Training-based methods~\cite{sd,sdedit} have first been proposed to solve this problem. For all time steps, they retrain the time-dependent conditional score estimator~\cite{controlnet} to measure the distance between the condition and each intermediate result. Meanwhile, retraining the conditional score estimator requires a large amount of computation.

Another line of research is training-free conditional diffusion models~\cite{dps,RED,freedom,manifold,MCGD}, which reuse the unconditional score function for the conditional score estimator and do not require retraining. To make the conditional score estimator not a time-dependent score function and training-free, one solution is posterior sampling~\cite{dps}. Specifically, given the $t$-th step intermediate result $x_t$ and the condition $c$, it uses two steps to make the conditional term not be a time-dependent function: 1) Posterior mean. It first approximates the posterior mean $\hat{x}_0$ from $x_t$ by using the unconditional score function; and 2) Measurement. It then uses the energy function~\cite{freedom} or other measures~\cite{RED} to learn the relationship between the time-independent $\hat{x}_0$ and the condition $c$. Hence, we can achieve a training-free method by reusing the unconditional score function in the first step and using the same measure for different time steps since the $\hat{x}_0$ is time-independent. Both are training-free. In this way, the conditional term will guide the intermediate results to be close to the condition. However, the iterative generation process still causes slow conditional generation.

Previous methods alleviated the time cost by introducing additional hypotheses. For example,
Gabriel et al.~\cite{MCGD} redefined the conditional term based on the Bayesian framework.
RED~\cite{RED} fine-tuned the start point of the reverse process based on variational inference.
Further, MPGD~\cite{manifold} introduced the linear-based manifold hypotheses to decouple the dependency between the conditional term and the diffusion model.

In this paper, we offer a novel view based on the Fisher information~\cite{information_thoertic} for the conditional score estimator, and a novel Fisher information-based conditional diffusion (FICD) is proposed to reduce the time cost while maintaining high-quality generation.

Concretely, according to the two steps, i.e., the posterior mean and measurement, in posterior sampling for training-free methods, we also divide the conditional score estimator into 1) 
the posterior part (more details in Sec.~\ref{sec:method}) and 2) the measurement part (more details in Eq.~\ref{eq:partial}). Then, we find that the posterior part can be redefined as the Fisher information, where the Fisher information could be exactly reflected in the information gain. Based on this, the posterior part could be further regarded as the weight function for the measurement part, which measures how much information could be gained for the condition in each time step. Following this new view, we also find that the weight function may sometimes hinder the reverse process from being close to the condition. Therefore, this motivates us to use the upper bound of Fisher information to increase the overall information gain. In this way, with more information gained following the conditional, the reverse process could generate images closer to the condition. Meanwhile, the upper bound also helps us cancel the calculation of the diffusion model's gradients simultaneously. Therefore, our FICD could accelerate the conditional generation while maintaining high quality.
To sum up, the main contributions of this paper are:
\begin{itemize}
    \item We propose a novel FICD to decrease the computation cost but increase the generation quality, which uses the upper bound of the Fisher information to approximate the posterior mean by incorporating the information theory.
    
    \item The proposed method provides a new perspective to understand and improve the existing training-free conditional diffusion methods, where the key is to accumulate enough information gain for the condition in the reverse process.
    
    \item The experimental results show that FICD accelerates the generation speed while maintaining high quality compared with SOTA methods.
\end{itemize}

\section{Related Work}
\textbf{Training-based conditional diffusion models}. These methods aim to fine-tune the parameters of the score estimator for different downstream tasks. For example, DreamBooth~\cite{dreambooth} directly fine-tuned the UNet of the diffusion model for the condition. ControlNet~\cite{controlNet++} introduced an additional UNet and fine-tuned it while freezing the original one. Stable Diffusion~\cite{sd} introduced the transformer layers to re-train. Training-based methods could generate high-quality images according to the condition, but the cost is too high since it needs to be fine-tuned each time for different downstream tasks.

\textbf{Training-free conditional diffusion methods}. Some training-free methods focus on utilizing the structure and potential of the UNet. For example, Tumanyan \textit{et al.}~\cite{pnp} leveraged the attention maps from attention layers. Jaeseok \textit{et al.}~\cite{hspace} found the ``h-space" among the UNet of the diffusion model. Wu \textit{et al.}~\cite{cyclediffusion} directly changed the inputs of the UNet. Except for the success, these methods need a special design based on the downstream tasks, thus limiting their generalization.

The other methods~\cite{dps,pse-guidance,freedom,freetuner} focus on sampling from the posterior distribution. For example, DPS~\cite{dps} first used the distance norm as the metric based on the inverse problem. FreeDom~\cite{freedom} further used the energy function as the metric, which extended it from solving the inverse problem to more widely downstream tasks such as face ID generation~\cite{reconstruction}. Yong-Hyun~\cite{RiemannianGeometry} leveraged the bias vector discovered in the latent space to guide the diffusion models. For the inverse problem, RED~\cite{RED} and Gabriel~\cite{MCGD} fine-tuned this start point and rebuilt the reverse process based on the Bayesian framework to improve the generation quality and decrease the time cost. DSG~\cite{DSG} alleviated the estimation bias of the condition in the reverse process to improve the FreeDom and mainly for the inverse problem.

Recently, MPGD~\cite{manifold} has been proposed to cancel the posterior part based on the manifold assumption. Compared to the MPGD, FICD offers an additional view based on the information theory to accelerate the sampling process while maintaining high-quality generation. MPGD introduced the manifold hypothesis based on linear assumption to decouple the dependency, while our method has no additional assumption. Meanwhile, it could be found that dropping the score function's gradient will inevitably lose useful information, which is different from the MPGD and may lead to an unstable generation for some tasks.

\textbf{Information theory with the diffusion models}. In light of the SDE~\cite{sde} that introduced the score function to explain the behavior of DMs, information theory~\cite{information_thoertic} shows the potential to make further improvements to DMs. Both information theoretical diffusion~\cite{information_thoertic} and InforDiffusion~\cite{infodiffusion} leveraged mutual information to interpret the correlation behind the observed and hidden variables to improve the generation quality. Interpretable diffusion~\cite{interpretable_mutual} further explained the most important part of the input for DMs similar to CAM~\cite{cam}.

\section{Preliminary}
\label{sec:pre}
\textbf{Training-free conditional generation.} Training-free methods aim to solve conditional generation tasks without retraining the unconditional diffusion model. Suppose that we have an unconditional diffusion model, the unconditional score function is $\nabla_{\bm{x}_{t}}\log p(\bm{x}_{t})$ in the $t$-th timestep~\cite{ddpm}. To keep the consistency with SDE~\cite{sde}, we also denote $s_{\theta}(\bm{x}_{t},t) \approx \nabla_{\bm{x}_{t}}\log p(\bm{x}_{t})$, where $s_{\theta}(\bm{x}_{t},t)$ is the score function based on the neural network with $\theta$ parameters. Now given the condition $\bm{c}$, the conditional score function can be formulated as $\nabla_{\bm{x}_{t}}\log p(\bm{x}_{t}|\bm{c})$. The problem becomes how to define this conditional score estimator.

By Bayesian rule~\cite{dps}, we have:
\begin{equation}
    \nabla_{\bm{x}_{t}}\log p(\bm{x}_{t}|\bm{c}) =  \underbrace{\nabla_{\bm{x}_{t}}\log p(\bm{x}_{t})}_{\text{Unconditional term}} + \underbrace{\nabla_{\bm{x}_{t}}\log p(\bm{c}|\bm{x}_{t})}_{\text{Conditional term}}.
    \label{eq:bayesian}
\end{equation}
The unconditional term is the unconditional score function $s_{\theta}(\bm{x}_{t},t)$.  To estimate the conditional term, a differentiable metric $\varepsilon(\bm{x}_{t},\bm{c})$ (also called the energy function~\cite{freedom}) is proposed to measure the distance between $\bm{x}_{t}$ and $\bm{c}$:
\begin{equation}
    p(\bm{c}|\bm{x}_{t}) = \frac{\exp^{-\lambda \varepsilon(\bm{x}_{t},\bm{c})}}{Z},
    \label{eq:energy function}
\end{equation}
where $\lambda$ is a temperature coefficient and $Z>0$ is a normalizing constant.

Following the posterior sampling~\cite{freedom}, we calculate the posterior mean $\bm{\hat{x}}_{0|t}$ of the $\bm{x}_{t}$:
\begin{equation}
        \bm{\hat{x}}_{0|t} \approx \frac{1}{\sqrt{\hat{\alpha_{t}}}}(\bm{x}_{t} + (1-\hat{\alpha}_{t})s_{\theta}(\bm{x}_{t},t)),
    \label{eq:mmse2}
\end{equation}
where $\hat{\alpha}_{t}$ is also a known parameter from the noise schedule~\cite{ddpm} related to the $t$-th timestep. 
And then we use  $\bm{\hat{x}}_{0|t}$ to measure the distance between the condition $\bm{c}$ under the data domain instead of the noise domain~\cite{dps}. Based on Eq.~\ref{eq:energy function} and ~\cite{dps}, we have:
\begin{equation}
        \log p(\bm{c}|\bm{x}_{t}) \approx  \log p(\bm{c}|\bm{\hat{x}}_{0|t}) \propto \varepsilon(\bm{\hat{x}}_{0|t},\bm{c}).
    \label{eq:mmse}
\end{equation}

\textbf{Motivation}. Based on the Eq.~\ref{eq:bayesian}-Eq.~\ref{eq:mmse}, we take the derivative of the $\log p(\bm{c}|\bm{x}_{t})$ with respect to $\bm{x}_{t}$ and have
\begin{equation}
    \nabla_{\bm{x}_{t}}\log p(\bm{c}|\bm{\hat{x}}_{0|t}) = \frac{\partial \bm{\hat{x}}_{0|t}}{\partial \bm{x}_{t}}\frac{\partial \varepsilon(\bm{\hat{x}}_{0|t},\bm{c}) }{\partial \bm{\hat{x}}_{0|t}}.
    \label{eq:partial}
\end{equation}
According to Eq.~\ref{eq:partial}, the derivative of the conditional term with respect to $\bm{x}_{t}$ in training-free methods can be further divided into two parts. The first part is the derivative of the posterior mean $\bm{\hat{x}}_{0|t}$ with respect to $\bm{x}_{t}$: $\frac{\partial \bm{\hat{x}}_{0|t}}{\partial \bm{x}_{t}}$, called it \textit{posterior part}. This part does not contain the condition $\bm{c}$. The second part is the derivative of the measurement function with respect to the posterior mean: $\frac{\partial \varepsilon(\bm{\hat{x}}_{0|t},\bm{c}) }{\partial \bm{\hat{x}}_{0|t}}$. We refer to it as the \textit{measurement part}, which contains the condition $\bm{c}$. 

In this paper, we show that the Fisher information could explain the posterior part as the information gain. Following this, the posterior part is similar to acts as the weight function for the measurement part to control the accumulated information about how $x_{t}$ is close to the condition $c$. In this case, we further show an interesting finding: we could increase the information gain of the posterior part to increase the generation quality. To achieve this, we propose using the Fisher information's upper bound as the new posterior part, where the upper bound could further reduce the time cost.

\section{Methodology}
\label{sec:method}

In our work, we introduce the Fisher information to cancel the posterior part $\frac{\partial \bm{\hat{x}}_{0|t}}{\partial \bm{x}_{t}}$. Specifically, we leverage the upper bound of the Fisher information to redefine the posterior part. Additionally, we provide an additional view from the information theory to explain how the Fisher information improves the training-free conditional generation.

\textbf{Fisher information-based conditional diffusion.} We expand the posterior part as follows:
\begin{equation}
\begin{split}
        \frac{\partial \bm{\hat{x}}_{0|t}}{\partial \bm{x}_{t}} = \frac{1}{\sqrt{\hat{\alpha}_{t}}}(1 + (1-\hat{\alpha}_{t})\frac{\partial s_{\theta}(\bm{x}_{t},t)}{\partial \bm{x}_{t}}).
\end{split}
    \label{eq:partial extention}
\end{equation}
The intractable part is $\frac{\partial s_{\theta}(\bm{x}_{t},t)}{\partial \bm{x}_{t}}$.In this condition, we introduce the Fisher information to help us investigate $\frac{\partial s_{\theta}(\bm{x}_{t},t)}{\partial \bm{x}_{t}}$. We have the following definition:
\begin{definition}
The gradient of the unconditional score function with respect to $\bm{x}_{t}$ is precisely the definition of the Fisher information~\cite{InformationTheory}. 
\begin{equation}
    I(\bm{x}_{t}) = \frac{\partial s_{\theta}(\bm{x}_{t},t)}{\partial \bm{x}_{t}},
    \label{eq:fisher information}
\end{equation}
where $I(\bm{x}_{t})$ is the Fisher information related to the variable $\bm{x}_{t}$, which measures the information that $\bm{x}_{t}$ carried out in the $t$-th time.
\label{th:fisher_information_defineation}
\end{definition}
The detailed proofs are provided in the supplementary. 

\begin{figure}[h!]
     \centering
    \begin{subfigure}[b]{0.22\textwidth}
    \centering
       \includegraphics[width=\textwidth]{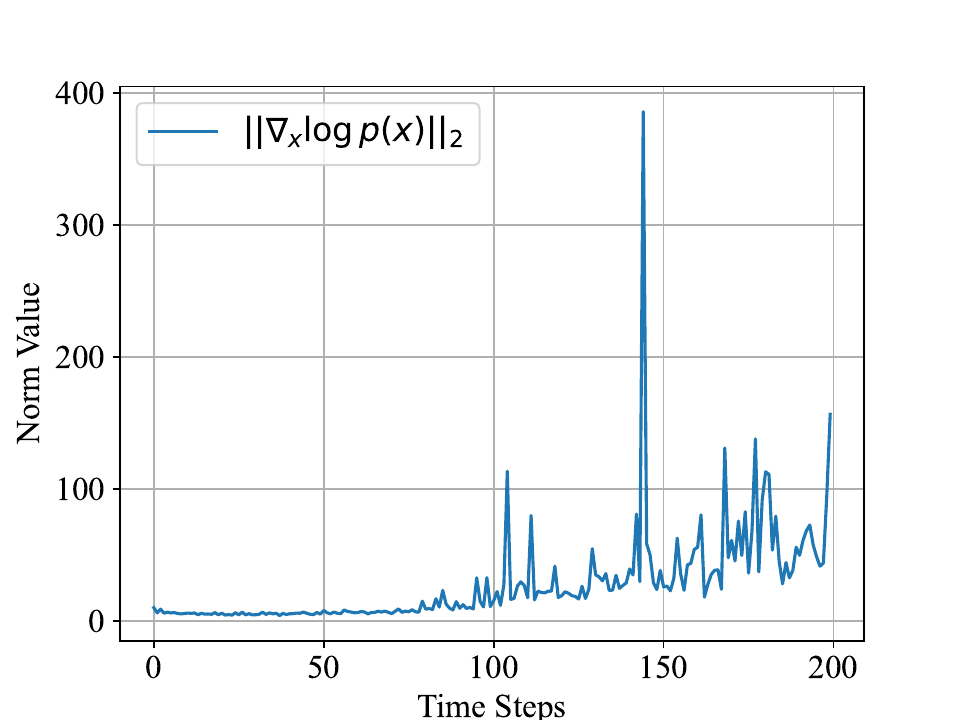}
       \caption{$T=200$}
    \end{subfigure}
    \begin{subfigure}[b]{0.22\textwidth}
        \centering
       \includegraphics[width=\textwidth]{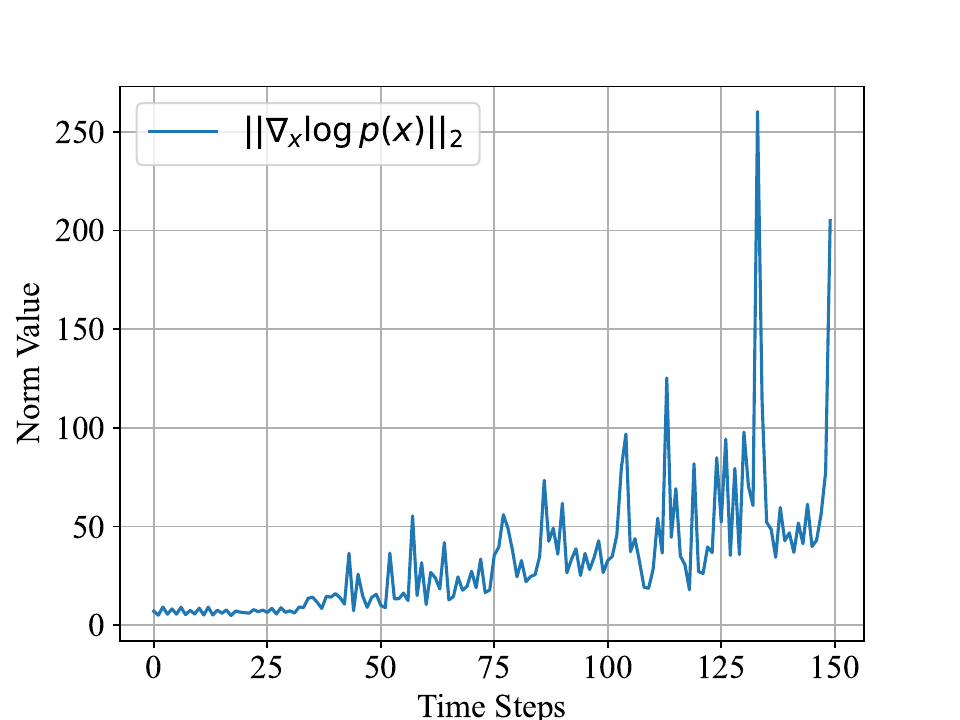}
       \caption{$T=150$}
    \end{subfigure}
    \begin{subfigure}[b]{0.22\textwidth}
        \centering
       \includegraphics[width=\textwidth]{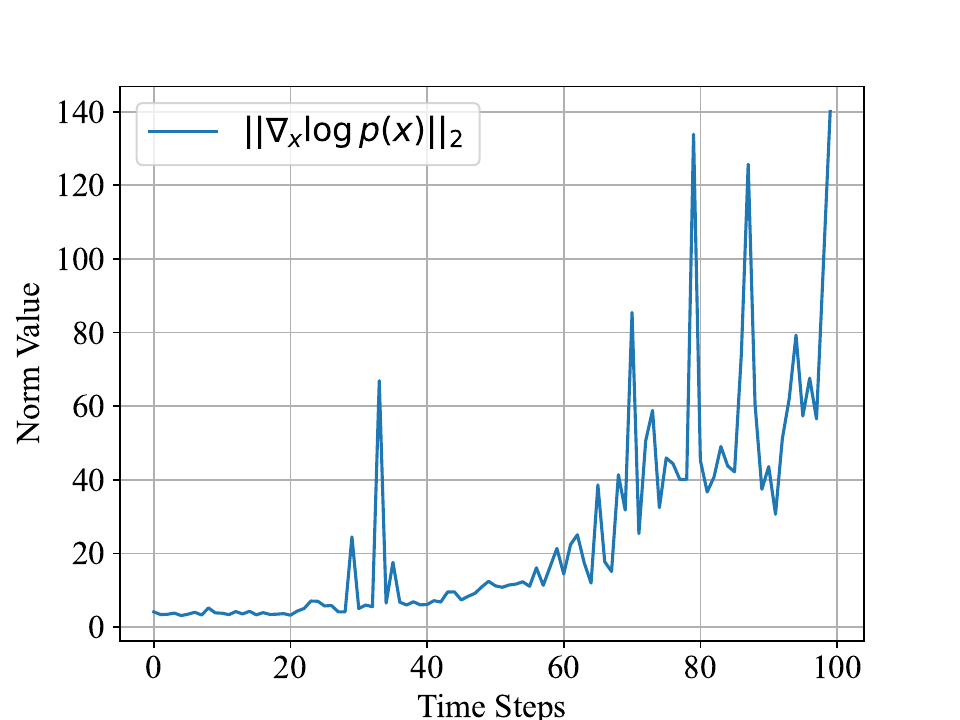}
       \caption{$T=100$}
    \end{subfigure}
    \begin{subfigure}[b]{0.22\textwidth}
        \centering
       \includegraphics[width=\textwidth]{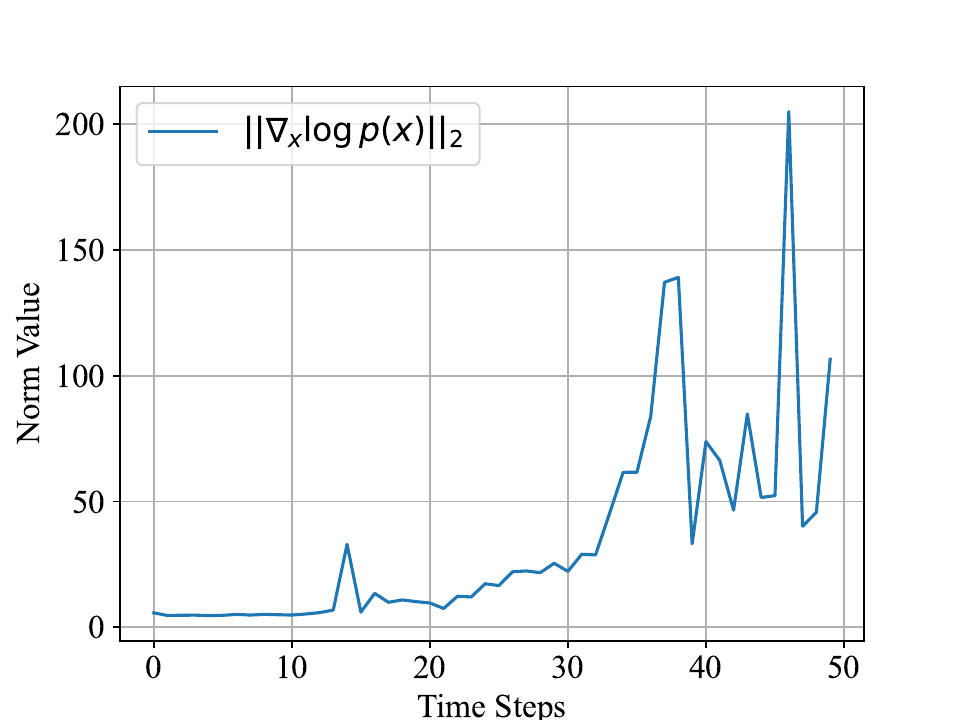}
       \caption{$T=50$}
    \end{subfigure}
    \caption{An empirical study for gradient norm based on the style guidance tasks with stable diffusion. We show the value of the $||\nabla_{\bm{x}_{t}}\log p(\bm{c}|\bm{\hat{x}}_{0|t})||_2$ among different timesteps to show the information gain. Concretely (a), (b), (c), and (d) report the values under 200, 100, 50, and 30 sampling steps respectively.} 
    \label{fig:overall}
\end{figure}

\textbf{Cram$\acute{e}$r-Rao bound estimation}. Interestingly, we could leverage the upper bound of $I(\bm{x}_{t})$, the Cram$\acute{e}$r-Rao bound estimation to cancel the posterior part:
\begin{theorem}
Given the sequence $\{\bm{x}_{T},\bm{x}_{T-1},...,\bm{x}_{t},...,\bm{x}_{1}\}$, where $t \in [T, 0)$ and $\bm{x}_{T}$ is the initial state of the reverse process, the $I(\bm{x}_{t})$ is bounded to the Cram$\acute{e}$r-Rao bound:
    \begin{equation}
        I(\bm{x}_{t}) <  \frac{1}{1-\hat{\alpha}_{t}}.
        \label{eq:upperbound}
    \end{equation}
    \label{th:upperbound}
\label{th:convergent theory}
\end{theorem}

\begin{algorithm}[t]
    \caption{The overall algorithm for FICD} \label{al:stage1}
    \label{al:FIGD}
    \begin{algorithmic}[1]
     \Statex \textbf{Input:} $\bm{c}$, $T$, $s_{\theta}$, the noise schedule parameter $\hat{\alpha}_{t}$, $\beta_{t}$, the differentiable metric function $\varepsilon$ and the hyperparameter $\rho_{t}$.
     \Statex \textbf{Output:} $x_{0}$ \Comment{The generated image based on $c$}
    \State $x_{T}\sim \mathcal{N}(0,1)$
    \For{$t$ in $[T-1,...,1]$} 
        \State $\epsilon \sim \mathcal{N}(0,1)$
        \State $\bm{x}_{t-1}= (1+0.5\beta_{t})\bm{x}_{t} +  \beta_{t}\nabla_{\bm{x}_{t}}\log p(\bm{x}_{t}) + \sqrt{\beta_{t}}\epsilon$
        \State $\bm{\hat{x}}_{0|t}=\frac{1}{\sqrt{\hat{\alpha}_{t}}}(\bm{x}_{t} + (1-\hat{\alpha}_{t})s_{\theta}(\bm{x}_{t},t))$ \Comment{The MMSE estimation}
        \State $g_{t} = \frac{2}{\sqrt{\hat{a}_{t}}}\nabla_{\bm{\hat{x}}_{0|t}}\varepsilon(\bm{\hat{x}_{0|t}},\bm{c})$
        \State $\bm{x}_{t-1} = \bm{x}_{t-1} - \rho_{t}g_{t}$
    \EndFor \label{ref:end}
    \Statex\textbf{Return:} $x_{0}$
    \end{algorithmic}
\end{algorithm}

We replace the $I(\bm{x}_{t})$ by Cram$\acute{e}$r-Rao bound directly. In the end, the Eq.~\ref{eq:partial extention} could be estimated as:
\begin{equation}
    \frac{\partial \bm{\hat{x}}_{0|t}}{\partial \bm{x}_{t}} \approx \frac{2}{\sqrt{\hat{a}_{t}}}.
    \label{eq:estimation gradient}
\end{equation}
Eq.~\ref{eq:estimation gradient} shows the upper bound for $I(\bm{x}_{t})$ cancel the posterior part based on the noise scheduler parameters.

\textbf{Sampling process for FICD}. In the end, we could derive a new sampling process for FICD. By Eq.~\ref{eq:partial} and Eq.~\ref{eq:estimation gradient}, we have:
\begin{equation}
\begin{split}
        \nabla_{\bm{x}_{t}}\log p(\bm{c}|\bm{\hat{x}}_{0|t}) &\approx \frac{2}{\sqrt{\hat{a}_{t}}}\frac{\partial \varepsilon(\bm{\hat{x}}_{0|t},\bm{c}) }{\partial \bm{\hat{x}}_{0|t}} \\
        &=  \frac{2}{\sqrt{\hat{a}_{t}}}\nabla_{\bm{\hat{x}}_{0|t}}\log p(\bm{c}|\bm{\hat{x}}_{0|t}).
        \label{eq:estimate conditional generation}
\end{split}
\end{equation}
\begin{figure*}
    \centering
    \includegraphics[width=0.9\textwidth]{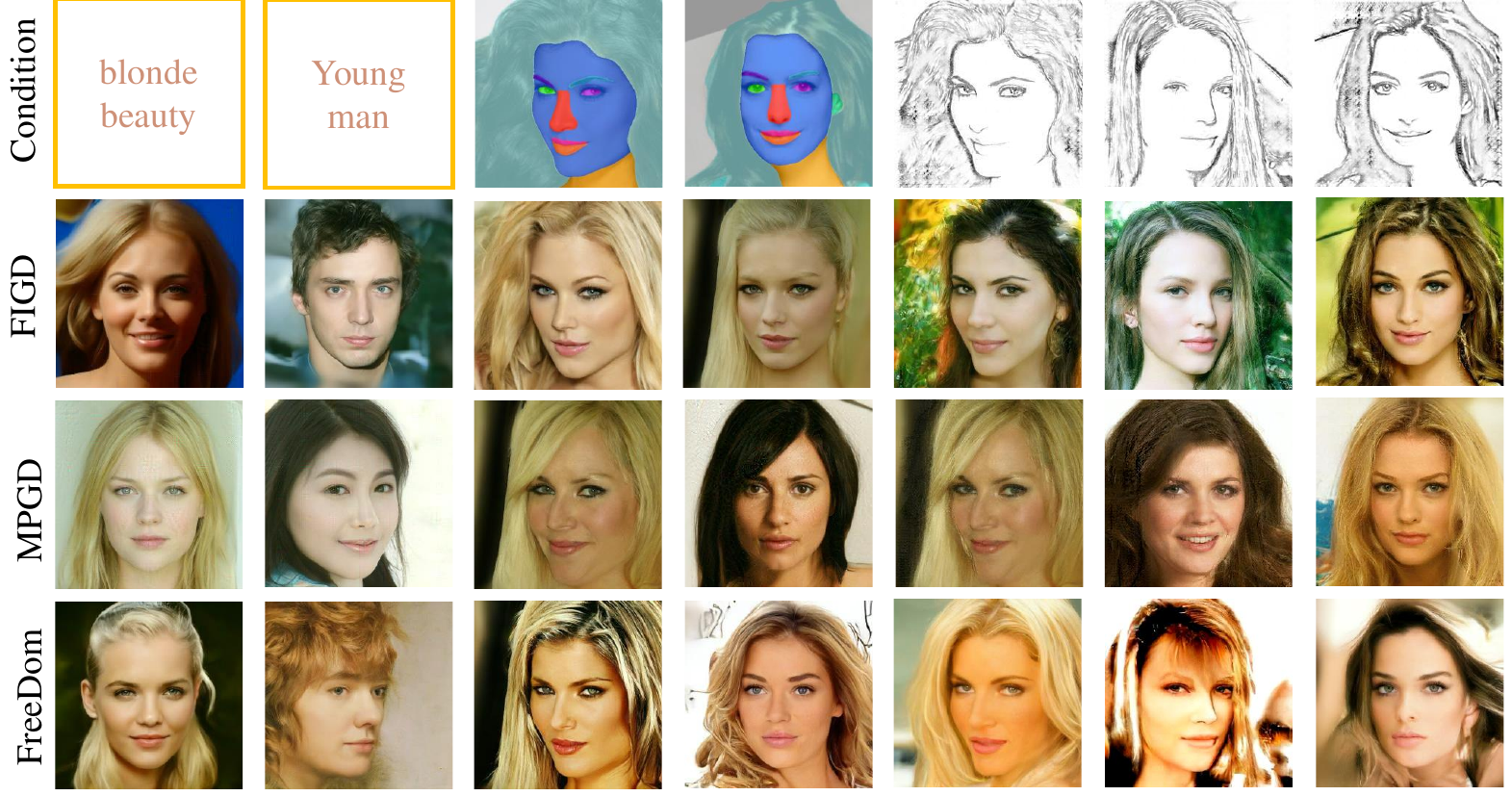}
    \caption{Qualitative examples of using a single condition human face images. The included conditions are (a) text, (b) face parsing maps, and (c) sketches. We compare the results with those of three baselines. It can be found that MPGD is invalid since these tasks break the linear hypothesis theory, and FICD performs well.}
    \label{fig:faceid}
\end{figure*}
Then by Eq.~\ref{eq:bayesian} and Eq.~\ref{eq:estimate conditional generation}, the sampling process of FICD is:
\begin{equation}
\begin{gathered}
    \bm{x}_{t-1} = \hat{m}_{t-1} - \frac{2\rho_{t}}{\sqrt{\hat{a}_{t}}}\nabla_{\hat{x}_{0|t}}\varepsilon(\bm{\hat{x}}_{0|t},\bm{c}) \\
    \hat{m}_{t-1} = (1+0.5\beta_{t})\bm{x}_{t} +  \beta_{t}\nabla_{\bm{x}_{t}}\log p(\bm{x}_{t}) + \sqrt{\beta_{t}}\epsilon,
    \label{eq:final sampling}
\end{gathered}
\end{equation}
where $\rho_{t}$ is the hyperparameter regarded as the learning rate to control the strength of guidance. The detailed algorithm is shown in Algorithm.~\ref{al:FIGD}. 

\subsection{An explanation via information-based perspective}
We offer an additional explanation to illustrate why our method can perform better than the existing training-free methods. Our explanation is based on the information perspective. Definition~\ref{th:fisher_information_defineation} illustrates that the posterior part could be regarded as information gain. This helps us formulate an information-based perspective view. 

Concretely, the measure part offers the direction following the condition. The overall $\frac{\partial \bm{\hat{x}}_{0|t}}{\partial \bm{x}_{t}}\frac{\partial \varepsilon(\bm{\hat{x}}_{0|t},\bm{c}) }{\partial \bm{\hat{x}}_{0|t}}$ could further regard the accumulated information following the direction of the condition. Its norm $|\nabla_{\bm{x}_{t}}\log p(\bm{c}|\bm{\hat{x}}_{0|t})|_{2}$ denotes to reflect the amount of the information gain in the $t$ time steps.
It could be noticed that the larger norm $|\nabla_{\bm{x}_{t}}\log p(\bm{c}|\bm{\hat{x}}_{0|t})|_{2}$ means more information.

\begin{figure}
     \centering
    \begin{subfigure}[b]{0.22\textwidth}
    \centering
       \includegraphics[width=\textwidth]{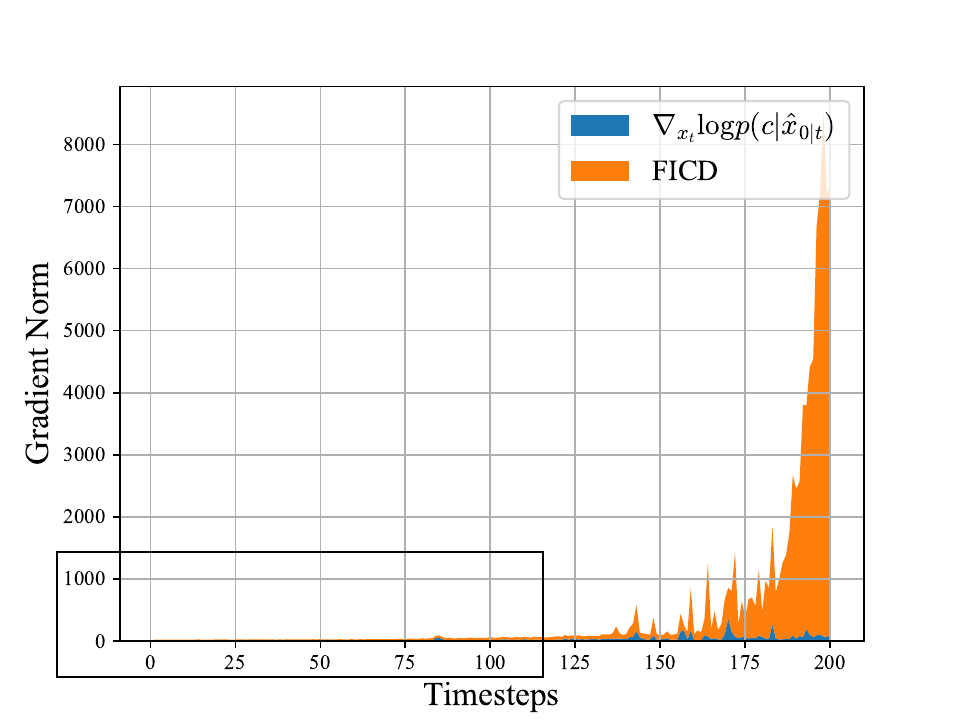}
       \caption{$T=200$}
    \end{subfigure}
    \begin{subfigure}[b]{0.22\textwidth}
        \centering
       \includegraphics[width=\textwidth]{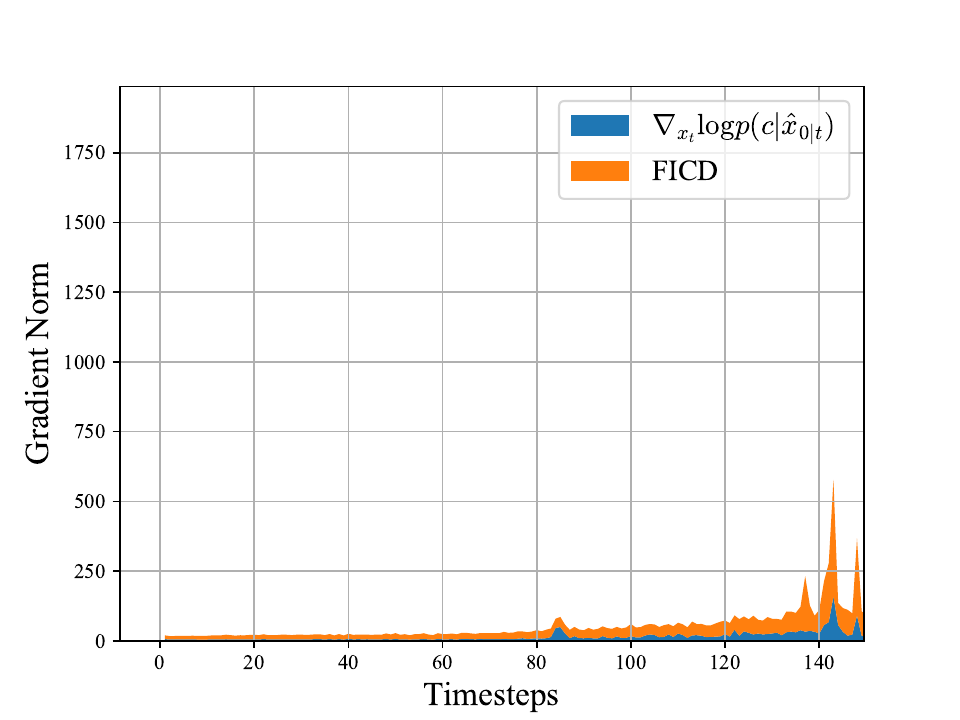}
       \caption{$T=100$}
    \end{subfigure}
    \begin{subfigure}[b]{0.22\textwidth}
        \centering
       \includegraphics[width=\textwidth]{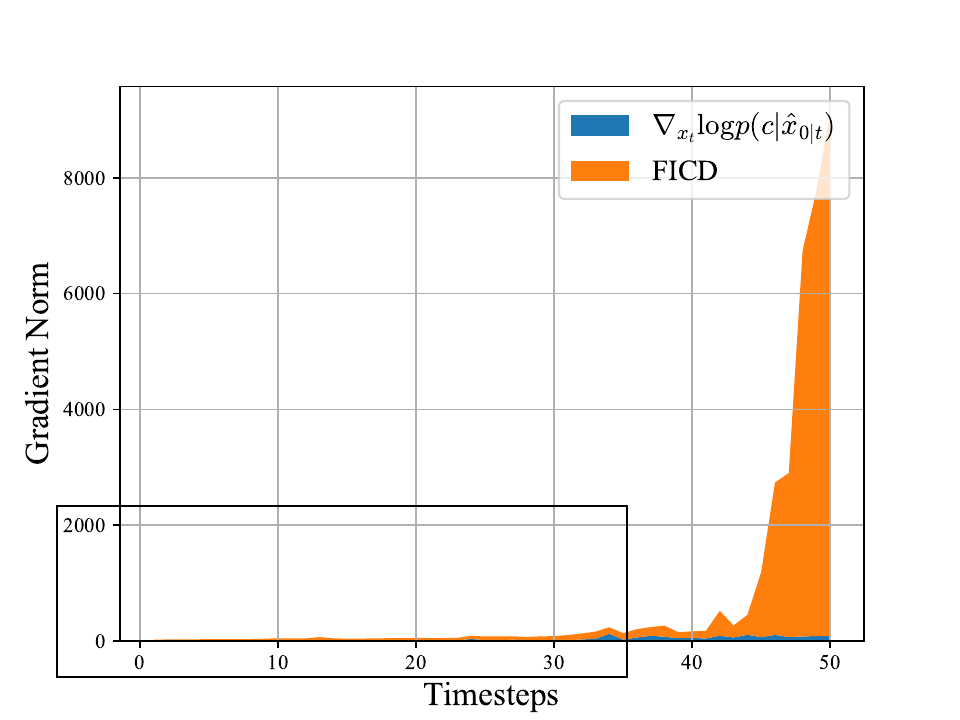}
       \caption{$T=50$}
    \end{subfigure}
    \begin{subfigure}[b]{0.22\textwidth}
        \centering
       \includegraphics[width=\textwidth]{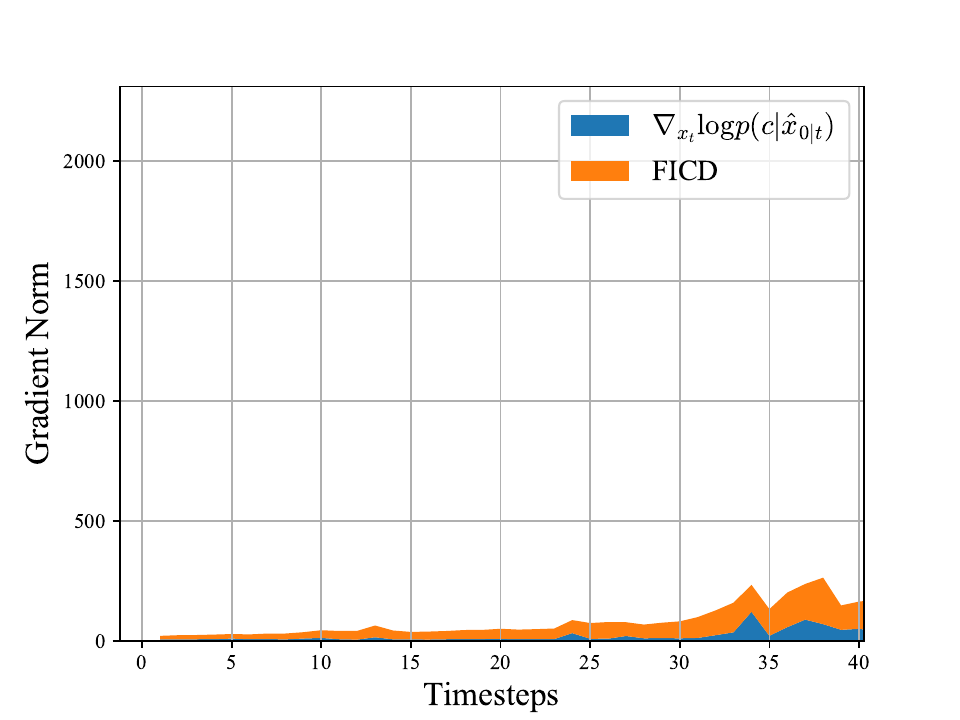}
       \caption{$T=30$}
    \end{subfigure}

    \caption{The comparison between FICD and $||\nabla_{\bm{x}_{t}}\log p(\bm{c}|\bm{\hat{x}}_{0|t})||_2$. (a) and (c) shows the value of the gradient norm between $\nabla_{\bm{x}_{t}}\log p(\bm{c}|\bm{\hat{x}}_{0|t})$ and FICD under $T=200$ and $T=50$ respectively. (b) and (d) is the sub-view for (a) and (c) repetitively. } 
    \label{fig:FIGD}
\end{figure}

First, for the overall conditional term $\nabla_{\bm{x}_{t}}\log p(\bm{c}|\bm{\hat{x}}_{0|t})$, some recent work~\cite{freedom} had shown that the early and latter phases could be skipped, and the critical phase is the middle phase in the conditional term. To further explore it, we also make an empirical study based on the style-generation task shown in Fig.~\ref{fig:FIGD} via showing the $|\nabla_{\bm{x}_{t}}\log p(\bm{c}|\bm{\hat{x}}_{0|t})|_{2}$ in different phases.  As shown in Fig.~\ref{fig:FIGD}, two observations can be concluded: 1) There is less information in the early and later phases since $|\nabla_{\bm{x}_{t}}\log p(\bm{c}|\bm{\hat{x}}_{0|t})|_{2}$ is smaller. Small $|\nabla_{\bm{x}_{t}}\log p(\bm{c}|\bm{\hat{x}}_{0|t})|_{2}$ generates a small weight for the conditional term to optimize the Eq.~\ref{eq:energy function}. 2) The key for conditional generation is the middle phase, since $|\nabla_{\bm{x}_{t}}\log p(\bm{c}|\bm{\hat{x}}_{0|t})|_{2}$ keeps in a suitable level. Different timesteps $T$ show a similar trend. In this information-based perspective, in the early and late phases, the existing methods~\cite{dps} do not generate samples well toward the condition, where a lot of time steps are wasted.

Intuitively, more information should be accumulated in both the early and late phases, especially for the early phase, where the distance between $x_{t}$ in the early phase and $c$ is very large. But, the above empirical results show that the existing methods do not provide enough information for the conditions.
Furthermore, we show that the upper bound of the proposed method can help us increase the information about the condition, especially in the early phase. Please note that after reformulating the posterior part by the upper bound, since $1- \hat{a}_{t}$ will tend to be small, thus $\frac{2}{\sqrt{\hat{a}_{t}}}$ now tend to large. In this condition, more information could be gained in the early phase, which generates a significant weight from the posterior part for the measure part. The early phase could contribute more to optimizing the Eq.~\ref{eq:energy function}. In this way, FICD could generate the conditional sample in the early phase, which reduces the time cost. We also report the empirical studies shown in Fig.~\ref{fig:FIGD} to show that the upper bound could increase the weight of the measure part in the early phase, which proves our explanation.

\section{Experiment}
\begin{figure*}
    \centering
    \includegraphics[width=\textwidth]{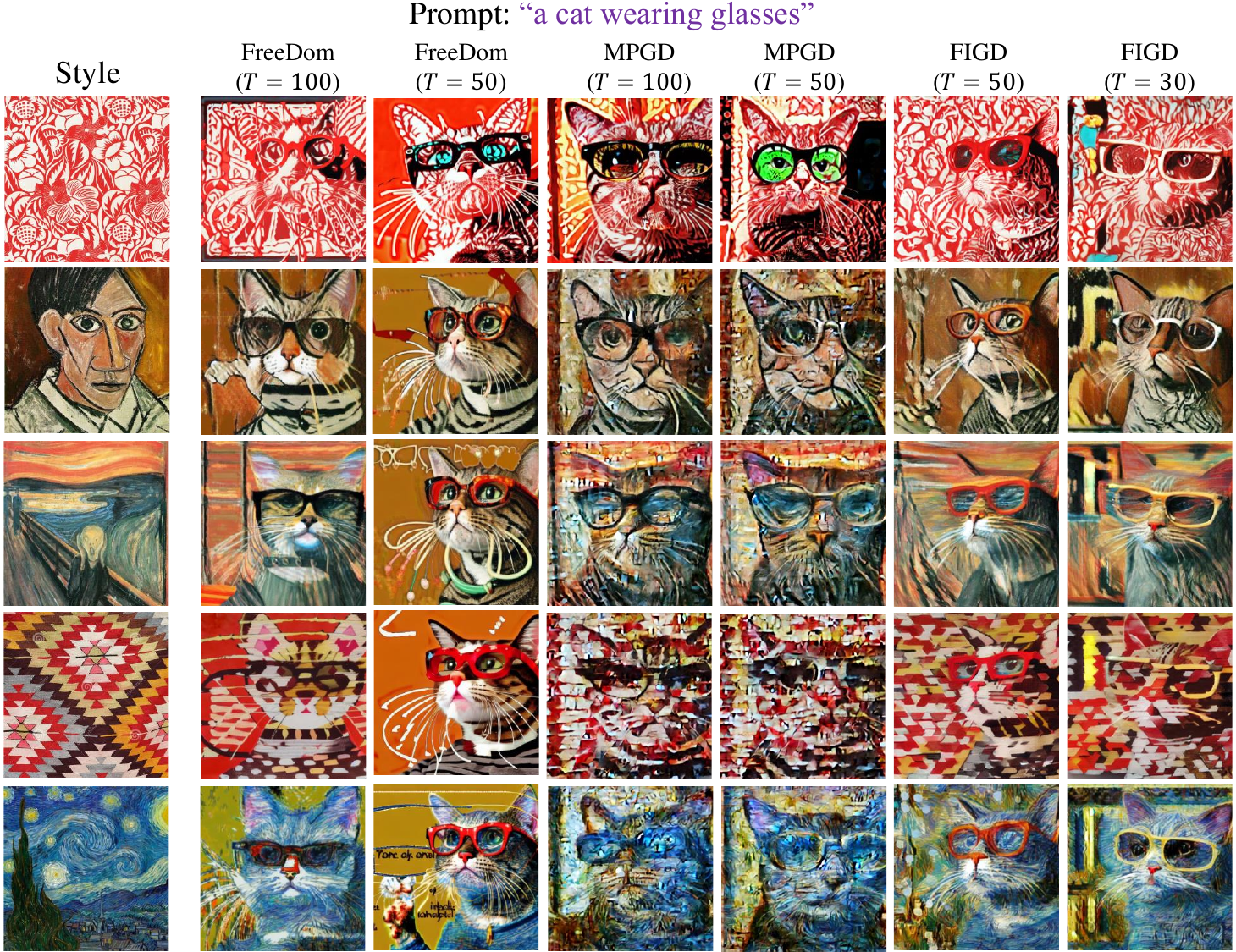}
    \caption{Qualitative examples of style-guided generation with Stable Diffusion experiment based on FICD compared with the three baselines.}
    \label{fig:style}
\end{figure*}
\subsection{Experiment Settings}
To demonstrate the potential of FICD, we focus on tasks in which the differentiable metrics are nonlinear based on various open-source diffusion models. The tasks in this paper mainly contain \textit{face-related tasks}, \textit{style-guided generation with stable diffusion}, and \textit{ControlNet-related generations with multiple guidance}. 

For the face-related tasks, we introduce text, segmentation maps, and sketches as the condition to guide the generation process following the FreeDom~\cite{freedom}. For the style-guided generation, we introduce a style image as the condition to guide the generation process. For the ControlNet-related generations, we introduce complex multi-guidance. The detailed settings, including the measurement for these tasks and the hyperparameters we have used, are listed in the supplementary. 

We use a single RTX4090 GPU to finish all the experiments. The baselines we used in our paper are three SOTA methods: \textit{FreeDom}, \textit{MPGD}, and \textit{LGD}~\cite{lgd}. We use the time-travel strategy~\cite{freedom}. To make a fair comparison, all the pre-trained diffusion models we used are the same as the FreeDom.

\begin{table}[htb]
    \caption{Quantitative results of style-guided generation with Stable Diffusion experiment. We compared FICD with the three baselines, FreeDom, MPGD, and LGD, in style-guided generation tasks with the Stable Diffusion; we get the overall improvement both in the Style score~\cite{freedom}, CLIP score, and the time cost since our method could generate high-quality images with few sampling steps. NA means that we used the shared memory since the requirement GPU VRAM of LGD-MC is over 24GB, and we cannot get the precise time cost.}
    \label{tab:style}
    \centering
    \begin{tabular}{c c c c}
        \toprule
        \textbf{Method} &  \textbf{Style$\downarrow$} &  \textbf{CLIP$\uparrow$} & \textbf{Time(s)} \\
        \midrule
        LGD-MC ($T=100$) & 247.39 & \textbf{29.96} & NA \\
        FreeDom ($T=100$) & 226.26 & 28.07 & 63 \\
        MPGD ($T=100$) & 351.16 & 25.08 & 27 \\
        FICD ($T=100$)& \textbf{225.83} & 29.59 & 27\\
        \midrule
        MPGD ($T=50$) & 408.28 & 23.23 & 17\\
        FreeDom ($T=50$) & 377.96 & 30.01 & 33 \\ 
        FICD ($T=50$)& \textbf{245.51} & \textbf{30.32} & 17\\
        FICD ($T=30$) & 281.67 & 29.07 & \textbf{11}\\
        \bottomrule
    \end{tabular}
\end{table}

\begin{table}
    \caption{Qualitative results of face-related tasks based on ControlNet experiments, where Pose distance is the norm between pose maps and generated images.}
    \label{tab:face_control}
    \centering
    \begin{tabular}{c c c c}
    \toprule
         \textbf{Method} &  \textbf{Pose Distance}$\downarrow$ & \textbf{Time(s)}  \\
    \midrule
         FreeDom &        54.21        &   193     \\
         FICD &          \textbf{40.01}     & \textbf{42}     \\
    \bottomrule
    \end{tabular}
\end{table}
\begin{figure*}
    \centering
    \includegraphics[width=\textwidth]{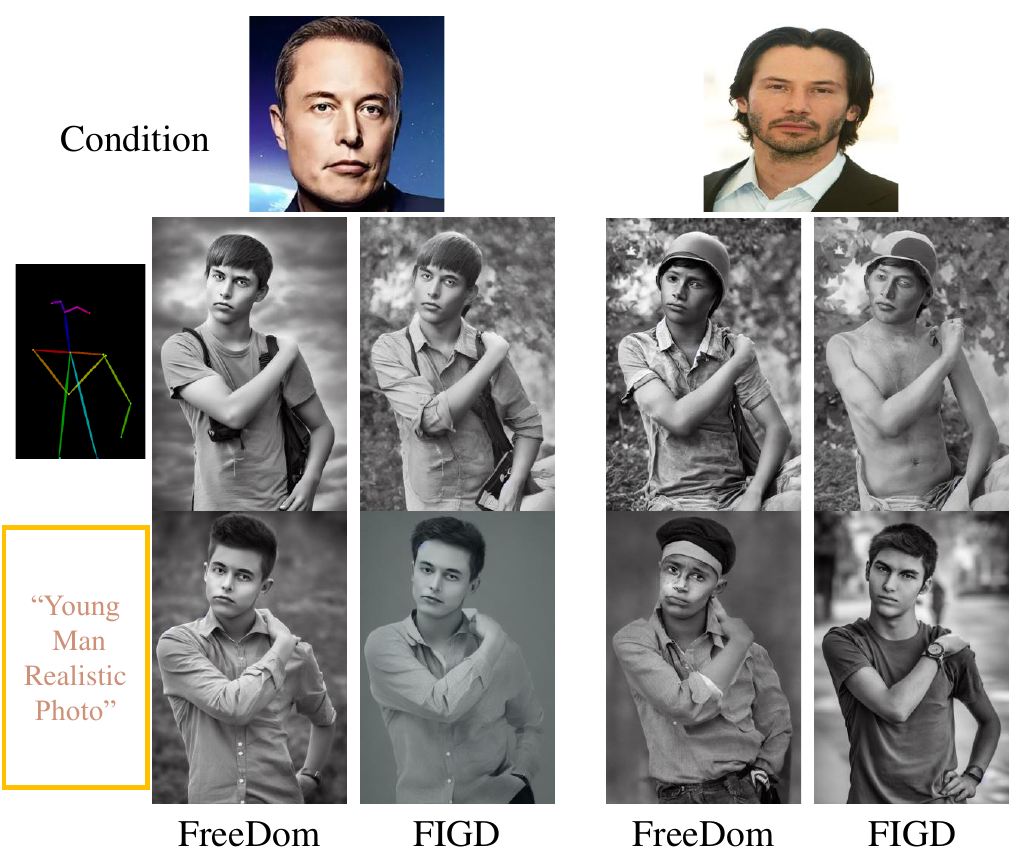}
    \caption{Qualitative examples of face-related tasks based on ControlNet experiments. We compared FICD with FreeDom and MPGD.}
    \label{fig:controlnetface}
\end{figure*}
\subsection{Qualitative Results}
\textbf{Face-related tasks}. To begin with, we first show the result of FICD in face-related tasks, including text, face parsing maps, and sketches, as shown in Fig~\ref{fig:faceid}. It could be found that FICD generates high-quality images related to the condition compared with the MPGD. Meanwhile, compared with the FreeDom, FICD generates similar images. This proves the validity of our theory analysis, where FICD increases the information to generate high-quality images further. We also report the qualitative results in the supplementary.

\textbf{Style-guided generation with Stable Diffusion.} To further show the potential of the FICD, we change the tasks to the style transfer based on the diffusion model. This task is more complicated than face-related tasks since there are two guides: 1) the text prompts and 2) the style image. We report the qualitative examples shown in Fig.~\ref{fig:style} and the qualitative results in Table~\ref{tab:style}. It can be seen that FICD the SOTA performance. Under the $T=100$, FICD achieves 225.83 Style metric while maintaining the 29.59 CLIP, which shows a SOTA trade-off compared to the other baselines. Meanwhile, with the decrease in the time step from $T=50$ to $T=30$, FICD achieved the SOTA results. These results show that improving the information gain could enhance efficiency while verifying that the conflict may lead to insufficient accumulated information.

\begin{figure*}
    \centering
    \includegraphics[width=\textwidth]{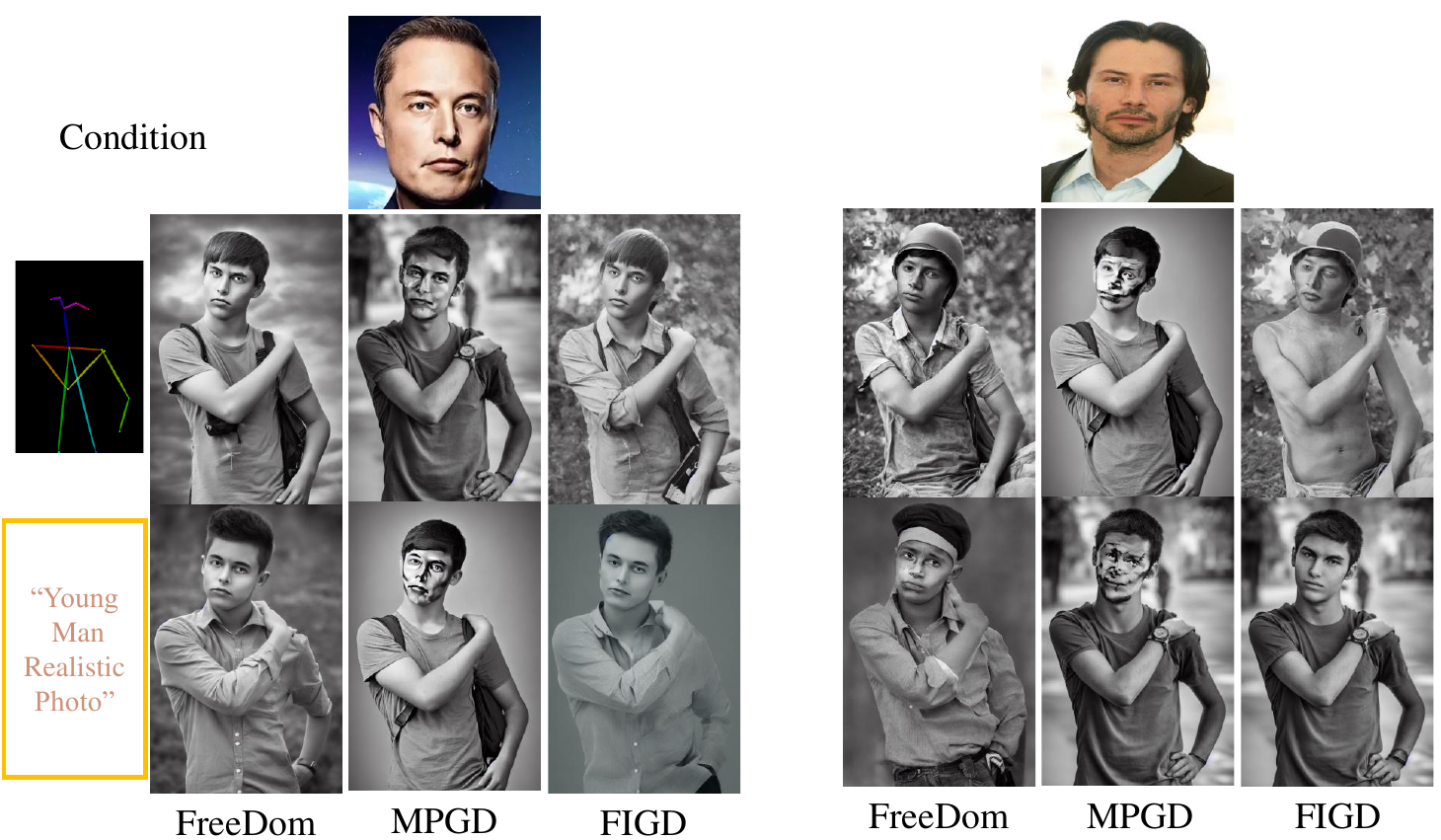}
    \caption{Qualitative examples of the style task based on ControlNet experiments. We compared FICD with the FreeDom. The text prompts are ``bike" for a sketch bike.}
    \label{fig:controlnetstyle}
\end{figure*}

\textbf{ControlNet-related generations with multiple guidance.} To further show the effectiveness of the FICD, following the FreeDom, we implement it in the multiple guidance tasks based on the ControlNet, which includes two tasks: 1) \textit{face guidance} and \textit{style} tasks. Concretely, for the face guidance tasks, we use the text prompts and pose mappings as the input of ControlNet to generate images with similar poses and satisfy the description of text prompts. In this condition, we add face ID images as the independent condition to guide the ControlNet in generating similar faces and poses that satisfy the text prompt and face ID. Then, We report the qualitative examples and the qualitative results shown in Fig~\ref{fig:controlnetface}. Since the MPGD will generate obvious mismatch generation based on the qualitative examples, we further compare the FreeDom shown in Table~\ref{tab:face_control}. It can be noticed that FICD improves the pose distance metric from 54.21 to 40.01 compared to the FreeDom. Then, the time cost reduces from the 193s to 42s. This illustrates the efficiency of the FICD, which further verifies the validity of the Fisher information.

For the style task, we changed the condition of the ControlNet to the sketch and used the style image as independent guidance to prove the feasibility of the FICD. We report the qualitative examples shown in Fig.~\ref{fig:controlnetstyle}. It can be noticed that FreeDom works well compared to the MPGD. Thus, we make a further comparison between FreeDom and FICD shown in Table~\ref{tab:controlnet_style}. It can be further found that FICD improves the CLIP metric from 69.71 to 70.27. Then, FICD improves the Style metric from 282.25 to 244.50. The time cost has been reduced from 299s to 32s by FICD. These results show the efficiency of the FICD. Meanwhile, it also proves that the conflict will truly cause the accumulated information not to be enough, and improving the information gained could enhance the generation quality.

\begin{table}
    \caption{Qualitative results of style image guidance with ControlNet experiments, where CLIP is the cosine similarity of the clip embedding of generated images and sketch, Style is the distance norm between style images and generated images.}
    \label{tab:controlnet_style}
    \centering
    \begin{tabular}{c c c c c}
    \toprule
         \textbf{Method}  & \textbf{CLIP}$\uparrow$ & \textbf{Style}$\downarrow$ & \textbf{Time(s)}  \\
    \midrule
         FreeDom &    69.71      & 282.25&299 \\
         FICD &      \textbf{70.27}   & \textbf{244.50}& \textbf{32}\\
    \bottomrule
    \end{tabular}
\end{table}

\textbf{Ablation study for $\rho_{t}$.} We study the effect of $\rho_{t}$ from small to large shown ($\rho_{t} =1$ as the beginning) in the supplementary. We can see that FICD is scalable, and the user can set different values according to the requirements.

\section{Conclusion}
We proposed the Fisher information conditional diffusion to achieve training-free condition generation called FICD. FICD first finds that the conditional term under the training-free conditional generation could be split into the measurement and posterior parts. Then, the posterior part could be modeled using Fisher's information. In this novel view, the function of the conditional term could be explained as accumulating enough information following the measurement part. The posterior part plays a role in controlling how much information can be accumulated. In this case, the novel insight is to use the upper bound of the Fisher information to approximate the posterior part. This leads to an increase in the overall information gain in the generation process. An interesting finding emerges here: increasing the information gain could enable FICD to achieve conditional generation with fewer steps, thus first reducing the time cost. This also improves the generation quality since more information can be accumulated during the overall generation process. To prove our theory, we offer an information theory-based explanation, giving an illustration based on the information theory to show why FICD could work well. In the end, the experimental results show that FICD could successfully increase the generation quality while reducing the time cost.

\textbf{Limitation.} However, there are some limitations. We have shown that dropping the gradient will cause some information to be lost. No theory identifies whether such information always has a negative influence. Thus, this will inevitably cause a potential threat. For example, based on further exploration, we find that dropping the gradient will inevitably aggregate conflict among various conditions. This is obvious in the ControlNet-related tasks. This is another reason why our upper bound could work better than only increasing the factor of the condition (Detailed in supplementary). In future work, we will further study to improve FICD.

%Bibliography
\bibliographystyle{unsrt}  
\bibliography{main}  

\begin{thebibliography}{10}

\bibitem{ddpm}
Jonathan Ho, Ajay Jain, and Pieter Abbeel.
\newblock Denoising diffusion probabilistic models.
\newblock {\em CoRR}, abs/2006.11239, 2020.

\bibitem{sde}
Yang Song, Jascha Sohl-Dickstein, Diederik~P Kingma, Abhishek Kumar, Stefano Ermon, and Ben Poole.
\newblock Score-based generative modeling through stochastic differential equations.
\newblock In {\em International Conference on Learning Representations}, 2021.

\bibitem{ddim}
Jiaming Song, Chenlin Meng, and Stefano Ermon.
\newblock Denoising diffusion implicit models, 2022.

\bibitem{controlnet}
Lvmin Zhang, Anyi Rao, and Maneesh Agrawala.
\newblock Adding conditional control to text-to-image diffusion models, 2023.

\bibitem{sd}
Robin Rombach, Andreas Blattmann, Dominik Lorenz, Patrick Esser, and Björn Ommer.
\newblock High-resolution image synthesis with latent diffusion models, 2021.

\bibitem{glide}
Alex Nichol, Prafulla Dhariwal, Aditya Ramesh, Pranav Shyam, Pamela Mishkin, Bob McGrew, Ilya Sutskever, and Mark Chen.
\newblock Glide: Towards photorealistic image generation and editing with text-guided diffusion models, 2022.

\bibitem{endtoend}
Bram Wallace, Akash Gokul, Stefano Ermon, and Nikhil Naik.
\newblock End-to-end diffusion latent optimization improves classifier guidance, 2023.

\bibitem{pnp}
Narek Tumanyan, Michal Geyer, Shai Bagon, and Tali Dekel.
\newblock Plug-and-play diffusion features for text-driven image-to-image translation.
\newblock In {\em Proceedings of the IEEE/CVF Conference on Computer Vision and Pattern Recognition (CVPR)}, pages 1921--1930, June 2023.

\bibitem{freedom}
Jiwen Yu, Yinhuai Wang, Chen Zhao, Bernard Ghanem, and Jian Zhang.
\newblock Freedom: Training-free energy-guided conditional diffusion model.
\newblock {\em Proceedings of the IEEE/CVF International Conference on Computer Vision (ICCV)}, 2023.

\bibitem{sdedit}
Chenlin Meng, Yutong He, Yang Song, Jiaming Song, Jiajun Wu, Jun-Yan Zhu, and Stefano Ermon.
\newblock Sdedit: Guided image synthesis and editing with stochastic differential equations, 2022.

\bibitem{dreambooth}
Nataniel Ruiz, Yuanzhen Li, Varun Jampani, Yael Pritch, Michael Rubinstein, and Kfir Aberman.
\newblock Dreambooth: Fine tuning text-to-image diffusion models for subject-driven generation.
\newblock In {\em {IEEE/CVF} Conference on Computer Vision and Pattern Recognition, {CVPR} 2023, Vancouver, BC, Canada, June 17-24, 2023}, pages 22500--22510, 2023.

\bibitem{dps}
Hyungjin Chung, Jeongsol Kim, Michael~Thompson Mccann, Marc~Louis Klasky, and Jong~Chul Ye.
\newblock Diffusion posterior sampling for general noisy inverse problems.
\newblock In {\em The Eleventh International Conference on Learning Representations}, 2023.

\bibitem{diifusionbeatgan}
Prafulla Dhariwal and Alex Nichol.
\newblock Diffusion models beat gans on image synthesis.
\newblock {\em CoRR}, abs/2105.05233, 2021.

\bibitem{RED}
Morteza Mardani, Jiaming Song, Jan Kautz, and Arash Vahdat.
\newblock A variational perspective on solving inverse problems with diffusion models, 2023.

\bibitem{manifold}
Yutong He, Naoki Murata, Chieh-Hsin Lai, Yuhta Takida, Toshimitsu Uesaka, Dongjun Kim, Wei-Hsiang Liao, Yuki Mitsufuji, J.~Zico Kolter, Ruslan Salakhutdinov, and Stefano Ermon.
\newblock Manifold preserving guided diffusion, 2023.

\bibitem{MCGD}
Gabriel Cardoso, Yazid Janati~El Idrissi, Sylvain~Le Corff, and Eric Moulines.
\newblock Monte carlo guided diffusion for bayesian linear inverse problems, 2023.

\bibitem{information_thoertic}
Xianghao Kong, Rob Brekelmans, and Greg {Ver Steeg}.
\newblock Information-theoretic diffusion.
\newblock In {\em International Conference on Learning Representations}, 2023.

\bibitem{controlNet++}
Ming Li, Taojiannan Yang, Huafeng Kuang, Jie Wu, Zhaoning Wang, Xuefeng Xiao, and Chen Chen.
\newblock Controlnet++: Improving conditional controls with efficient consistency feedback, 2024.

\bibitem{hspace}
Jaeseok Jeong, Mingi Kwon, and Youngjung Uh.
\newblock Training-free content injection using h-space in diffusion models, 2024.

\bibitem{cyclediffusion}
Chen~Henry Wu and Fernando~De la~Torre.
\newblock A latent space of stochastic diffusion models for zero-shot image editing and guidance.
\newblock In {\em ICCV}, 2023.

\bibitem{pse-guidance}
Jiaming Song, Arash Vahdat, Morteza Mardani, and Jan Kautz.
\newblock Pseudoinverse-guided diffusion models for inverse problems.
\newblock In {\em The Eleventh International Conference on Learning Representations, {ICLR} 2023, Kigali, Rwanda, May 1-5, 2023}, 2023.

\bibitem{freetuner}
Youcan Xu, Zhen Wang, Jun Xiao, Wei Liu, and Long Chen.
\newblock Freetuner: Any subject in any style with training-free diffusion, 2024.

\bibitem{reconstruction}
Zalan Fabian, Berk Tinaz, and Mahdi Soltanolkotabi.
\newblock Adapt and diffuse: Sample-adaptive reconstruction via latent diffusion models, 2023.

\bibitem{RiemannianGeometry}
Yong{-}Hyun Park, Mingi Kwon, Jaewoong Choi, Junghyo Jo, and Youngjung Uh.
\newblock Understanding the latent space of diffusion models through the lens of riemannian geometry.
\newblock In {\em Proceedings of the Advances in Neural Information Processing Systems 36: Annual Conference on Neural Information Processing Systems 2023, NeurIPS 2023, New Orleans, LA, USA, December 10 - 16, 2023}, 2023.

\bibitem{DSG}
Lingxiao Yang, Shutong Ding, Yifan Cai, Jingyi Yu, Jingya Wang, and Ye~Shi.
\newblock Guidance with spherical gaussian constraint for conditional diffusion, 2024.

\bibitem{infodiffusion}
Yingheng Wang, Yair Schiff, Aaron Gokaslan, Weishen Pan, Fei Wang, Christopher De~Sa, and Volodymyr Kuleshov.
\newblock {I}nfo{D}iffusion: Representation learning using information maximizing diffusion models.
\newblock In {\em Proceedings of the 40th International Conference on Machine Learning}, pages 36336--36354, 2023.

\bibitem{interpretable_mutual}
Xianghao Kong, Ollie Liu, Han Li, Dani Yogatama, and Greg~Ver Steeg.
\newblock Interpretable diffusion via information decomposition, 2023.

\bibitem{cam}
Ramprasaath~R. Selvaraju, Michael Cogswell, Abhishek Das, Ramakrishna Vedantam, Devi Parikh, and Dhruv Batra.
\newblock Grad-cam: Visual explanations from deep networks via gradient-based localization.
\newblock {\em International Journal of Computer Vision}, 128(2):336–359, October 2019.

\bibitem{InformationTheory}
Andrew~R. Barron.
\newblock Entropy and the central limit theorem.
\newblock {\em The Annals of Probability}, 14(1):336--342, 1986.

\bibitem{lgd}
Jiaming Song, Qinsheng Zhang, Hongxu Yin, Morteza Mardani, Ming{-}Yu Liu, Jan Kautz, Yongxin Chen, and Arash Vahdat.
\newblock Loss-guided diffusion models for plug-and-play controllable generation.
\newblock In {\em International Conference on Machine Learning, {ICML} 2023, 23-29 July 2023, Honolulu, Hawaii, {USA}}, pages 32483--32498, 2023.

\end{thebibliography}

\end{document}